\documentclass[journal]{IEEEtran}

\ifCLASSINFOpdf
\else
\usepackage[dvips]{graphicx}
\fi
\usepackage{url}

\usepackage{amsmath,amsfonts,bbm}
\usepackage{array}
\usepackage{textcomp}
\usepackage{stfloats}
\usepackage{url}
\usepackage{verbatim}
\usepackage{graphicx}
\usepackage{cite}

\usepackage{amssymb}
\usepackage{bm}
\usepackage{multirow}
\usepackage{booktabs}
\usepackage{lipsum}
\usepackage{enumerate}
\usepackage{multicol}
\usepackage{multirow}
\usepackage{subfigure}
\usepackage{stfloats}
\usepackage{bbding}

\usepackage[ruled,linesnumbered]{algorithm2e}

\usepackage[colorlinks,
linkcolor=blue,
anchorcolor=blue,
citecolor=blue]{hyperref}

\usepackage{halloweenmath}
\allowdisplaybreaks[4]

\usepackage{amsthm}
\usepackage{ulem}
\newtheorem{lemma}{Lemma}

\newtheorem{theorem}{Theorem}

\newtheorem{corollary}{Corollary}

\def\gap{0.75ex}
\abovedisplayskip\gap
\belowdisplayskip\gap
\abovedisplayshortskip\gap
\belowdisplayshortskip\gap
\begin{document}
	
\title{Mirror Online Conformal Prediction with Intermittent Feedback}

\author{Bowen Wang,~\IEEEmembership{Student Member,~IEEE}, 
Matteo Zecchin,~\IEEEmembership{Member,~IEEE}, and Osvaldo Simeone,~\IEEEmembership{Fellow,~IEEE}
\vspace{-2em}
\thanks{This work was partially supported by the European Union’s Horizon Europe project CENTRIC (101096379), by the Open Fellowships of the EPSRC (EP/W024101/1), and by the EPSRC project (EP/X011852/1).  The authors are with the King’s Communications, Learning and Information Processing (KCLIP) Lab, Department of Engineering, King’s College London, London WC2R 2LS, U.K. (e-mail: \{bowen.wang, matteo.1.zecchin, osvaldo.simeone\}@kcl.ac.uk).
}
}

%\markboth{Journal of \LaTeX\ Class Files, Vol. 14, No. 8, August 2015}
%{Shell \MakeLowercase{\textit{et al.}}: Bare Demo of IEEEtran.cls for IEEE Journals}
\maketitle

\begin{abstract}
Online conformal prediction enables the runtime calibration of a pre-trained artificial intelligence model using feedback on its performance. 
Calibration is achieved through set predictions that are updated via online rules so as to ensure long-term coverage guarantees. 
While recent research has demonstrated the benefits of incorporating prior knowledge into the calibration process, this has come at the cost of replacing coverage guarantees with less tangible regret guarantees based on the quantile loss. 
This work introduces intermittent mirror online conformal prediction (IM-OCP), a novel runtime calibration framework that integrates prior knowledge, operates under potentially intermittent feedback, and features minimal memory complexity.
IM-OCP guarantees long-term coverage and sub-linear regret, both of which hold deterministically for any given data sequence and in expectation with respect to the intermittent feedback.
\end{abstract}


\IEEEpeerreviewmaketitle

\vspace{-1em}
\section{Introduction}
\IEEEPARstart{I}{N} safety-critical and high-stakes sequential decision-making processes, such as in robotics \cite{kulic2005safe,lindemann2023safe}, wireless communications \cite{simeone2025conformal,zecchin2024forking}, finance \cite{lockwood2015predicting}, and medicine \cite{vazquez2022conformal,lu2022fair}, it is important to accurately quantify uncertainty in order to reliably predict the potential outcomes of given actions. 
Producing reliable uncertainty estimates in these scenarios is often challenging due to the non-stationarity of the data-generating distribution and to the possibility that feedback on the quality of the predictions may be only intermittently available \cite{mao2017survey}.

A formal approach to quantify uncertainty complements point predictions with prediction sets that satisfy coverage guarantees \cite{vovk2005algorithmic}. 
Notably, \textit{online conformal prediction} (OCP) \cite{angelopoulos2023conformal} schemes convert predictions produced by any black-box model into prediction sets that enjoy distribution-free, long-term coverage guarantees.
Specifically, \textit{adaptive conformal inference} (ACI) \cite{gibbs2021adaptive} adjusts prediction set sizes based on feedback about past decisions \cite{gibbs2021adaptive,feldman2023achieving,bhatnagar2023improved,angelopoulos2023pid,angelopoulos2024online,zecchin2024localized}. 
ACI can be wrapped around any predictive model, including classifiers \cite{xi2025robust}, regressors \cite{gibbs2021adaptive}, and sequential models (large language models) \cite{cherian2024large}.
Sparse feedback can be accommodated via \emph{intermittent ACI} (I-ACI)  \cite{zhao2024conformalized}.

In contrast to ACI, the recently introduced \emph{Bayesian ACI} (B-ACI) is an OCP method that adopts a data-centric Bayesian objective incorporating not only feedback, but also \emph{prior information} on the prediction scores  \cite{zhang2024benefit}. 
When the prior information is well specified, B-ACI can yield prediction sets with lower \emph{regret} in terms of the quantile loss as compared to ACI, which implies better performance in \textit{stochastic} settings with independent data \cite{zhang2024benefit}.
However, B-ACI does not guarantee \textit{deterministic} long-term coverage like ACI.
Furthermore, it requires storing past data, as well as solving a convex problem at each calibration step.

As summarized in Table \ref{tab:Com}, existing online calibration schemes either cannot incorporate prior information or sacrifice long-term coverage guarantees (see Sec. \ref{Sec:2} for details). 
This work addresses these limitations through the following key contributions:

\noindent $\bullet$ \textit{Intermittent mirror online conformal prediction (IM-OCP):} We introduce IM-OCP, a novel online calibration framework that builds on \textit{online mirror descent} (OMD) \cite{orabona2019modern, shalev2012online, hazan2016introduction} by integrating prior knowledge and incorporating an importance weighting strategy \cite{elvira2021advances} to handle intermittent feedback. 
IM-OCP exhibits constant memory usage and low-complexity updates.

\noindent $\bullet$ \textit{Regret and coverage guarantees:} We demonstrate that IM-OCP guarantees long-term coverage and sublinear regret, both of which hold deterministically for any data sequence and in expectation over the intermittent feedback.
While regret performance is basically inherited from OMD, establishing the coverage guarantees requires novel technical steps, including proving the boundedness of the IM-OCP iterates.

\noindent $\bullet$ \textit{Numerical validation:} We showcase the merits of IM-OCP for the task of \textit{received signal strength indicator} (RSSI)-based localization under intermittent feedback \cite{wu2012csi,Si2024IoTJ,Sifaou2025TCCN}.

\setlength{\tabcolsep}{4pt}
\begin{table}[]
\centering
\caption{Comparison with the State of the Art \vspace{-0.3cm} }
\begin{tabular}{@{}c|c|c|c|c|c@{}}
	\hline
	Methods                   & \begin{tabular}[c]{@{}c@{}}ACI\\ \cite{gibbs2021adaptive} \end{tabular}    & \begin{tabular}[c]{@{}c@{}}I-ACI \\ \cite{zhao2024conformalized} \end{tabular}        & \begin{tabular}[c]{@{}c@{}}B-ACI \\ \cite{zhang2024benefit} \end{tabular}        & \begin{tabular}[c]{@{}c@{}}IB-ACI  \end{tabular}       & \begin{tabular}[c]{@{}c@{}}IM-OCP \\ (this work) \end{tabular}       \\ \hline \hline
	long-term coverage     & \Checkmark & \Checkmark & \XSolidBrush     & \XSolidBrush     & \Checkmark \\ \hline
	sublinear regret       & \Checkmark & \Checkmark & \Checkmark & \Checkmark & \Checkmark \\ \hline
	prior information      & \XSolidBrush     & \XSolidBrush     & \Checkmark & \Checkmark & \Checkmark \\ \hline
	constant memory        & \Checkmark & \Checkmark & \XSolidBrush     & \XSolidBrush     & \Checkmark \\ \hline
	intermittent feedback  & \XSolidBrush     & \Checkmark & \XSolidBrush     & \Checkmark & \Checkmark \\ \hline
\end{tabular}
\label{tab:Com}
\vspace{-0.5cm}
\end{table}

\vspace{-1em}
\section{Background}\label{Sec:2}

\subsection{Online Calibration via Set Prediction}\label{sec:II-A}
We consider the problem of online calibration in sequential decision-making problems. 
At each round $t$, a learner observes an input $X_t \in \mathcal{X}$ and aims to predict the corresponding label $Y_t\in\mathcal{Y}$. 
To this end, the predictor assigns a negatively oriented \emph{score} $s(X_t, Y)$ to each value $Y\in\mathcal{Y}$, which is assumed to be upper bounded by a constant  $B < \infty$, i.e.,  $s(X,Y)\leq B$.
Using predictor's output, a \emph{prediction set} $\mathcal{C}_t$ is produced that includes all labels  $Y\in \mathcal{Y}$ with a score no larger than a \emph{threshold} $r_t$, i.e., \cite{angelopoulos2023conformal}
\begin{align}
\label{eq:pred_set}
\mathcal{C}_t = \mathcal{C}(X_t, r_t) = \left\lbrace y \in \mathcal{Y} : s(X_t, Y) \leq r_t \right\rbrace.
\end{align} 

After producing the decision $\mathcal{C}_t$, the system receives feedback in one of the following two forms, from least to most informative: 1) \emph{miscoverage error feedback}, in which the feedback signal at time $t$ corresponds to the miscoverage error $E_t=\mathbbm{1}\{Y_t \notin \mathcal{C}_t\}$; and 2) \emph{score feedback}, where the feedback corresponds to the ground truth score $r^*_t=s(X_t,Y_t)$. 
We assume that feedback is \emph{intermittent}, so that feedback is available with probability $p_t$ at round $t$.

Given a target miscoverage rate $\alpha \in [0, 1]$, the goal of calibration is to leverage the outlined feedback information to optimize, in an online fashion, the thresholds $\{r_t\}_{t=1}^T$ in (\ref{eq:pred_set}) so as to ensure the \emph{long-term deterministic coverage guarantee} over the time horizon $T$ \cite{gibbs2021adaptive} 
\begin{equation}\label{eq:coverage}
\mathrm{MisCov}(T) = \left| \frac{1}{T} \sum_{t=1}^T E_t - \alpha \right| \leq AT^{-\gamma} .
\end{equation}
In \eqref{eq:coverage},  the parameters $A$ and $\gamma$ are constants independent of $T$. By \eqref{eq:coverage}, 
the rate of miscoverage errors on an arbitrary sequence $\{(X_t, Y_t)\}_{t=1}^T$ converges to the target miscoverage level $\alpha$ as the number of steps $T$ tends to infinity.   

While the condition \eqref{eq:coverage} applies to an arbitrary sequence $\{(X_t, Y_t)\}_{t=1}^T$, in some settings one may be justified to assume that the sequence consists of \emph{independent and identically distributed}  (i.i.d.) samples. In this case,   the \emph{probabilistic}  coverage condition $\Pr\{Y_t \notin \mathcal{C}_t\}= \alpha$ can be met for each time $t$ by selecting the threshold $r_t$ as the $(1-\alpha)$-quantile of the score distribution. This quantile can be estimated by minimizing the empirical average of the \emph{quantile loss} \cite{feldman2023achieving,bhatnagar2023improved,angelopoulos2023pid,angelopoulos2024online,zecchin2024localized}
\begin{align}
\label{eq:pinball_loss}
\ell_{1-\alpha}(r, r^*_t) = (\alpha - \mathbbm{1}\{r < r^*_t\}) (r - r^*_t) ,
\end{align} 
as we have equality $q_{\alpha}(r^*_{1:T})=\min_{u \in \mathbb{R}} \sum_{t=1}^{T} \ell_{1-\alpha}(u, r^*_t)$. In fact, with i.i.d. data, the estimate $q_{\alpha}(r^*_{1:T})$ tends to the true $(1-\alpha)$-quantile  in the limit  $T\rightarrow \infty$.

To assess the capacity of a calibration procedure to perform well also in the case of i.i.d. data, it is thus useful to evaluate the extent to which the thresholds $\{r_t\}_{t=1}^T$ deviate from the estimated quantile $q_{\alpha}(r^*_{1:T})$. Note that the latter can be only evaluated in hindsight, i.e., at time $T$, while the thresholds $r_t$ must be produced online at time $t$. This deviation is measured by the \textit{regret} \cite{orabona2019modern, shalev2012online, hazan2016introduction}
\begin{align}
\label{eq:regret}
\mathrm{Reg}(T) = \sum_{t=1}^{T} \ell_{1-\alpha}(r_t, r^*_t) - \min_{u \in \mathbb{R}} \sum_{t=1}^{T} \ell_{1-\alpha}(u, r^*_t),
\end{align}
which amounts to the  difference between the \emph{cumulative quantile loss} of the predicted sequence $\{r_t\}_{t=1}^T$ and the cumulative loss of the optimal fixed threshold $q_{\alpha}(r^*_{1:T})$.

\vspace{-0.5em}
\subsection{Adaptive Conformal Inference}

Assuming that the feedback signal is always available, i.e.,  $p_t=1$ for all $t$, ACI \cite{gibbs2021adaptive} adaptively adjusts the thresholds $r_t$ using an \emph{online gradient descent} (OGD) strategy based on the miscoverage error feedback $E_t$. Specifically, the threshold  $r_t$ is updated using  the gradient of the quantile loss \eqref{eq:pinball_loss} as  
\begin{align}\label{eq:aci}
r_{t} = r_{t-1} - \eta_{t-1}(\alpha-E_{t-1}),
\end{align}  
where $\eta_t>0$ is the step size.  
By setting $\eta_t = c / \sqrt{t}$, ACI achieves a sublinear regret $\mathcal{O}(B\sqrt{T})$, while also ensuring the coverage guarantee in \eqref{eq:coverage} with $\gamma = 1/2$ \cite{angelopoulos2024online}.
Furthermore, given the single-pass nature of OGD \cite{orabona2019modern, shalev2012online, hazan2016introduction}, ACI enjoys constant memory complexity $\mathcal{O}(1)$.

ACI has been extended to the \emph{intermittent feedback}  scenario. Denoting as $\text{obs}_{t}$ the binary random variable taking value 1 if feedback $E_t$ is observed,  I-ACI \cite{zhao2024conformalized} uses an OGD  rule in which the step size is scaled as in 
\begin{equation}\label{eq:I-ACI-ODG}
r_t = r_{t-1} - \eta_{t-1}  (\alpha-E_{t-1}) \frac{\text{obs}_{t-1}}{p_{t-1}}.
\end{equation}  
I-ACI is shown to satisfy expected long-term coverage and to enjoy a sublinear expected regret (see \eqref{eq:new_mis} and \eqref{eq:Regret_IM_OCP}) \cite{zhao2024conformalized}.

\vspace{-1em}
\subsection{Bayesian Adaptive Conformal Inference}

ACI and I-ACI do not allow the integration of prior knowledge about the data distribution in the calibration process. In contrast,  B-ACI \cite{zhang2024benefit} assumes the availability of a prior distribution $P(r)$, with support $[0, B]$, on the scores $\{r^*_t\}_{t\geq 1}$. Leveraging this information, along with score feedback, at each time step $t$, the threshold $r_t$ is obtained by solving the convex optimization problem
\begin{equation}\label{eq:B_ACI}
r_t = \arg \min_{r} \left\{ h_{t} \psi(r) + \sum_{i=1}^{t-1} \ell_{1-\alpha}( r , r_i^* ) \right\},
\end{equation}  
where $h_{t} = \frac{\eta_{t} (t - 1)}{1 - \eta_{t}}$ and the convex regularizing function  $\psi(r) = \mathbb{E}_{r^* \sim P}[\ell_{1-\alpha}(r , r^*)]$ accounts for  prior knowledge.

B-ACI attains a sublinear regret $\mathcal{O}(D\sqrt{T})$, where the constant $D$ can be smaller than $B$ when the prior distribution is well specified \cite{zhang2024benefit}.
Thus, B-ACI has the potential to outperform ACI.
However, B-ACI is not guaranteed to satisfy the long-term coverage condition \eqref{eq:coverage}.
Furthermore, B-ACI requires a memory growing linearly over time as $\mathcal{O}(T)$ to store the sequence of feedback signals $\{r^*_t\}_{t \geq 1}$, and it necessitates solving the convex problem in \eqref{eq:B_ACI} at each round $t$.

Similar to ACI, it is possible to extend B-ACI to handle \emph{intermittent feedback}. This variant, referred to here as IB-ACI, obtains  the threshold $r_t$  by minimizing the sum of the importance-weighted losses
\begin{equation}\label{eq:IB-ACI}
r_t = \arg \min_{r} \left\{ h_{t} \psi(r) + \sum_{i=1}^{t-1} \ell_{1-\alpha}( r , r_i^* ) \frac{\text{obs}_i}{p_i} \right\}.
\end{equation}
The sublinear regret of IB-ACI can be easily established by leveraging standard results on the follow-the-regularized-leader \cite{orabona2019modern, shalev2012online, hazan2016introduction} method.

%\vspace{-1.em}
\section{Intermittent Mirror Online Conformal Prediction (IM-OCP)}
In this section, we present IM-OCP, a novel prior-dependent calibration scheme with intermittent feedback that exhibits low complexity, sublinear regret, and coverage guarantees. 

\vspace{-1.em}
\subsection{Algorithm Development}

IM-OCP incorporates prior information via a regularization term, and updates threshold $r_t$ via the application of OMD.

\subsubsection{Regularizer Encoding Prior Information}
For a given prior distribution $P(r)$ on the scores $r^*$ defined on the interval $[0, B]$, IM-OCP introduces a regularizer $R(\cdot)$ of the form  
\begin{equation}\label{eq:regularizer}
R(r) = \psi(r) + \frac{\sigma}{2} |r|^2 ,
\end{equation}  
where the first term follows B-ACI, while the second term ensures the strong convexity of the function $R(\cdot)$ for any fixed $\sigma > 0$.  
In Appendix \ref{appendix:B}, we show that the function $R(\cdot)$ is closed, differentiable, $\mu$-strongly convex for some real number $\mu>\sigma$, and $L$-smooth for some real number $L\le \max_{r \in [0, B]} \{P(r)\}+\mu$.

\subsubsection{Threshold Update via OMD}
Using the regularizer $R(r)$ in \eqref{eq:regularizer}, IM-OCP updates the threshold at each time $t$ according to the OMD rule \cite{orabona2019modern, shalev2012online, hazan2016introduction}.
To describe it, define the mirror map induced by function $R(r)$ as $\mathcal{M}_R(\cdot) = \nabla R(\cdot)$, and the corresponding inverse as $\mathcal{M}_R^{-1}(\cdot) = (\nabla R)^{-1}(\cdot)$.
The update rule of IM-OCP is then given as
\begin{align}
\label{eq:IM_OCP}
r_t = \mathcal{M}_R^{-1}\left(\mathcal{M}_R(r_{t-1}) - \eta_{t-1} (\alpha - E_{t-1}) \frac{\text{obs}_{t-1}}{p_{t-1}} \right).
\end{align}
When the feedback $E_{t-1}$ is observed ($\text{obs}_{t-1}$$=$$1$), this update corresponds to a gradient step in the dual space defined by function $R(\cdot)$. 
This adjusts the threshold $r_t$ in a direction informed by both the observed feedback and the prior knowledge embedded in the mirror map $\mathcal{M}_R(\cdot)$. 
In contrast, when no feedback is observed ($\text{obs}_{t-1}$$=$$0$), the update in \eqref{eq:IM_OCP} reduces to the identity map, and the threshold remains unchanged.

\begin{algorithm}[!t]
\SetAlgoLined
\SetAlgoVlined
\KwIn{Miscoverage rate $\alpha$, and prior information $P(r)$}
\textbf{Initialization:} $r_1 = 1-\alpha$ \\
\For{$t = 1 , \dots , T$}{
	\textbf{Observe} an \textit{input} $X_t \in \mathcal{X}$ \\
	Specify the \textit{score function} $s_t(X , Y)$ \\
	\textbf{Return} the \textit{prediction set} $\mathcal{C}(X_t , r_t)$ via eq. \eqref{eq:pred_set}\\
	Update the step size  $\eta_{t}$ based on Corollary \ref{coro:1}\\
	\If{$\textnormal{obs}_t = 1$}{
		Update threshold $r_{t+1}$ via eq. \eqref{eq:IM_OCP}\\
	}
	\Else{
		Keep threshold $r_{t+1} = r_t$ \\
	}
}
\caption{IM-OCP}
\label{alg:1}
\end{algorithm}

\subsubsection{Summary}
The proposed IM-OCP is summarized in Algorithm \ref{alg:1}. 
As shown, IM-OCP does not require storing past feedback values $Y_t^\star$, resulting in a constant memory. 
The inverse map $\mathcal{M}_R^{-1}(\cdot)$ can be evaluated offline using approximation methods such as neural network-based regression \cite{jun2020on}, entailing a minimal complexity and memory overhead.

\vspace{-1em}
\subsection{Theoretical Guarantees}\label{sec:3-A}

We now present the theoretical guarantees of IM-OCP. 
While the regret performance follows with minor changes from the standard theory of OMD \cite{orabona2019modern, shalev2012online, hazan2016introduction}, the analysis of long-term coverage is novel and challenging, requiring new proofs around the boundedness of the iterates produced by the update rule \eqref{eq:IM_OCP}.
Proofs are deferred to the \textcolor{blue}{supplementary material}. 

In the following, IM-OCP coverage is measured via its expected miscoverage error
\begin{equation}\label{eq:new_mis}
	\overline{\mathrm{MisCov}}(T) =   \left|  \mathbb{E} \left[\frac{1}{T}\sum_{t=1}^T E_t\right] - \alpha \right|,
\end{equation}
which corresponds to the absolute difference between the expected fraction of miscoverage errors, where the expectation is taken over the feedback observations $\{\text{obs}_t\}_{t=1}^T$, and the target coverage level $\alpha$.

We first establish the following ancillary result stating that the iterates produced by IM-OCP are bounded.
\vspace{-0.5em}
\begin{lemma}\label{pro:bound}
	For any $r_1\in[0,B]$, the iterates produced by IM-OCP satisfy
	the condition \begin{equation}
		r_t \in \left[ - \frac{\alpha \varpi_{t}}{\mu}  , B + \frac{(1 - \alpha) \varpi_{t}}{\mu} \right] \ \text{for $t>1$},
	\end{equation}
	where $\varpi_{t} = \max_{i \in [1, \dots, t-1]} \{ \eta_{i}/p_{i} \} $.
\end{lemma}
\vspace{-0.5em}

Equipped with Lemma \ref{pro:bound}, we characterize the expected miscoverage error rate of IM-OCP as follows.
\vspace{-0.5em}
\begin{theorem}\label{pro:mis_IMOCP}
	For any  non-increasing step size sequence $\{\eta_t\}_{t\geq 1}$, the miscoverage error rate of IM-OCP satisfies the inequality 
	\begin{equation}\label{eq:mis_cov_theorem}
		\overline{\mathrm{MisCov}}(T) \le  \frac{1}{T\eta_T} \left( L B + \frac{ L \eta_1 }{\mu p_{\min}}\right) ,
	\end{equation}
	where $p_{\min} = \min_{t \in [1,2,\cdots,T]} \{ p_t \}$.
\end{theorem}
\vspace{-0.5em}

IM-OCP also enjoys sublinear regret.
Similar to \eqref{eq:new_mis}, we define the IM-OCP regret as
\begin{equation}\label{eq:Regret_IM_OCP}
	\begin{aligned}
		& \overline{\mathrm{Reg}}(T)\hspace{-0.1em}=\mathbb{E} \left[ \sum_{i=1}^{T}{ \ell_\alpha ( r_i , r_i^* ) \frac{\text{obs}_i}{p_i} } \right]\hspace{-0.25em} - \min_{u \in \mathbb{R}}\sum_{i=1}^{T}\ell_\alpha (u , r_i^* ) ,
	\end{aligned}
\end{equation}
Like \eqref{eq:regret}, the quantity in \eqref{eq:Regret_IM_OCP} measures the difference between the expected cumulative quantile loss of IM-OCP and the loss of the best threshold chosen in hindsight having observed the entire sequence $\{r_t^*\}_{t\geq 1} $.
\vspace{-0.5em}
\begin{theorem}\label{pro:Regret_IM_OCP}
	For any non-increasing step size sequence $\{\eta_t\}_{t\geq 1}$, the regret of IM-OCP in \eqref{eq:Regret_IM_OCP} satisfies the inequality
	\begin{equation}\label{eq:new_reg}
		\begin{aligned}
			\overline{\mathrm{Reg}}(T) \le \frac{D_T}{\eta_T} + \frac{1}{2 \mu p_{\min}} \sum_{i=1}^T  \eta_i ,
		\end{aligned}
	\end{equation}
	where $D_T = \max_{t \in \{1,\dots, T\}} \mathcal{B}_R (q_{\alpha}(r^*_{1:T}) , r_t )$, with the $ \mathcal{B}_R(u,v)=R(u)-R(v)- (\nabla R(v))^T (u-v)$.
\end{theorem}
\vspace{-0.5em}

Theorems \ref{pro:mis_IMOCP} and \ref{pro:Regret_IM_OCP} provide bounds on the expected miscoverage error rate and regret of IM-OCP as a function of the learning rate sequence. 
The following corollary specializes the miscoverage and regret guarantee under a judicious choice of the learning rate.

\vspace{-0.5em}
\begin{corollary}\label{coro:1}  
	With a fixed learning rate $\eta = cT^{-\beta}$ or a decaying learning rate $\eta_t = ct^{-\beta}$, where $c$ is a positive constant independent of $T$, IM-OCP satisfies the miscoverage guarantee \eqref{eq:coverage} with $\gamma = 1-\beta$ and has regret $\mathrm{Reg}(T) = \mathcal{O} (T^{\max\{\beta,1-\beta\}})$.  
\end{corollary}
\vspace{-0.5em}

%\vspace{-1em}
\section{Numerical Results}
In this section, we present numerical results for an RSSI-based localization problem assuming intermittent feedback.

\vspace{-1em}
\subsection{Setting}
We consider a multi-building, multi-floor indoor localization problem and adopt the UJIIndoorLoc dataset \cite{torres2014ujiindoorloc}. The goal is to localize a device positioned in one of three buildings based on the signal power received from $M = 520$ access points \cite{wu2012csi,Si2024IoTJ,Sifaou2025TCCN}.  Accordingly, the input $X_t \in \mathbb{R}^M$ corresponds to an $M$-dimensional vector of RSSI values, while the label $Y_t = [Y_t^{\mathrm{Log}}, Y_t^{\mathrm{Lat}}]^T \in \mathbb{R}^2$ represents the longitude and latitude coordinates of the corresponding device to be localized. We apply online calibration to a pre-trained localization model $\hat{Y}_i = f_{\text{ELM}}(X_i)$, implemented using an extreme learning machine with a hidden layer of size 256 \cite{Sifaou2025TCCN}.
The model is trained using a training set consisting of 1000, 2000, and 8000 samples randomly selected from Buildings 1, 2, and 3, respectively. 
Since the accuracy of $f_{\text{ELM}}(\cdot)$ is expected to increase in the building index $i$, due to the increasingly larger datasets as $i$ grows, we set a decreasing probability of receiving feedback as $p_1 = 0.5$ for the first building, $p_2 = 0.3$ for the second, and $p_3 = 0.1$ for the third.
In these results, we focus on longitude prediction, using the prediction set $\mathcal{C}_t^\text{Lon}(X_t , r_t^\text{Lon} ) := \left[ [f_{\text{ELM}}(X_i)]_1 - r_t^\text{Lon} , [f_{\text{ELM}}(X_i)]_1 + r_t^\text{Lon} \right]$.
A similar approach can be applied to latitude prediction.

\vspace{-1em}
\subsection{Results}

We perform online calibration of the base predictor using  I-ACI, IB-ACI, and IM-OCP, considering both triangular and truncated Gaussian prior distributions $P(r)$ defined in the interval $[0, B]$.  
Specifically, for the triangular prior distribution, the mode is set to 0.1, while for the truncated Gaussian prior distribution, the mean is set to 0.1 and the variance to 2.
For all schemes, we consider a held-out data sequence of length $T = 2400$, with samples randomly selected from each building with equal probability, and set the target miscoverage rate to $\alpha=0.1$.

\begin{figure}[t]
	\centering
	\includegraphics[width=0.8\linewidth]{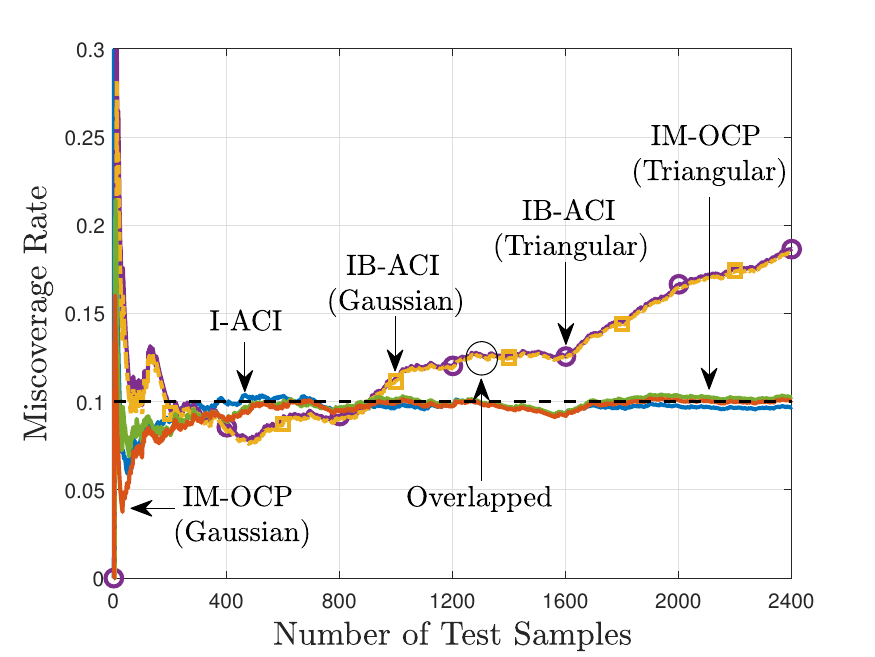}
	\vspace{-0.5em}
	\caption{Miscoverage rate versus the number of test samples ($\alpha = 0.1$).}
	\vspace{-0.5em}
	\label{fig:miscov}
\end{figure}

\begin{figure}[t]
	\centering
	\includegraphics[width=0.8\linewidth]{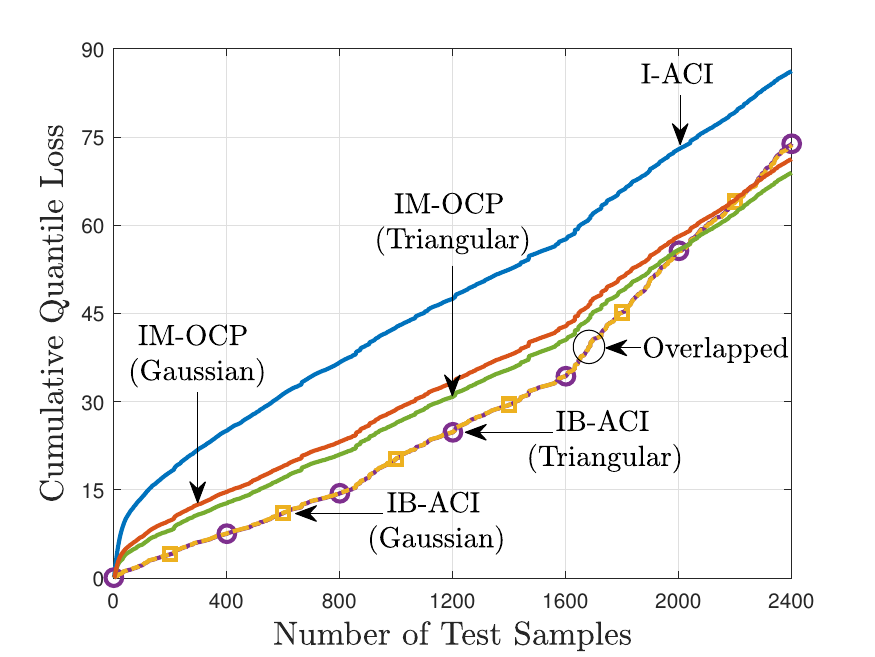}
	\vspace{-0.5em}
	\caption{Cumulative quantile loss versus the number of test samples ($\alpha = 0.1$).}
	\vspace{-0.5em}
	\label{fig:loss}
\end{figure}

In Fig. \ref{fig:miscov}, we present the miscoverage rate as a function of the time index $t$. 
Confirming the expected miscoverage guarantees \eqref{eq:mis_cov_theorem} provided in Theorem \ref{pro:mis_IMOCP}, the proposed IM-OCP exhibits a miscoverage rate that converges to the target level $\alpha$.
In contrast, IB-ACI, which does not come with a theoretical coverage guarantee, obtains a final miscoverage rate higher than the target value.

In Fig. \ref{fig:loss}, we report the cumulative quantile loss for all algorithms. 
By leveraging prior information, IM-OCP achieves a lower cumulative loss—and consequently, a smaller regret—compared to I-ACI, while remaining competitive with B-ACI.
Overall, these results verify the theoretical coverage guarantees of IM-OCP, while also demonstrating its capacity to incorporate prior information and achieve a lower regret compared to I-ACI.

\vspace{-0.5em}

\section{Conclusion}

We have proposed IM-OCP, an OCP scheme for intermittent feedback scenarios based on OMD, which uses importance sampling to handle intermittent feedback. 
IM-OCP allows the incorporation of prior knowledge about the calibration task via the specification of a mirror map, and is proven to achieve both sublinear regret and miscoverage guarantees. 
The practical effectiveness of IM-OCP is demonstrated through its application to an RSSI-based localization problem. 
Future work may extend this study to multi-agent online calibration, and explore online calibration under robustness requirements to impairments such as missing data.

\vspace{-0.5em}

\appendix

\subsection{Proof of Strongly Convex and Smoothness}\label{appendix:B}

The prior regularizer can be expressed as
\begin{align}
	& R(r) = \mathbb{E}_{r_i^* \sim P}[\ell_{1-\alpha}(r , r_i^*)] + \frac{\sigma}{2} |r|^2 \nonumber \\
	& = \int_{0}^{B} \ell_{1-\alpha}(r , r^*) P(r^*) \text{d}r^* + \frac{\sigma}{2} |r|^2 \\
	&  = \alpha \int_0^{B} (r - r^*) P(r^*) \text{d}r^* - \int_{r}^B (r - r^*) P(r^*) \text{d}r^* + \frac{\sigma}{2} |r|^2  . \nonumber
\end{align}
where $P(r)$ is the probability density functions of prior distribution $P$ in domain $[0, B]$.

The first derivative of $R(r)$ is
\begin{equation}
	\begin{aligned}
		& \nabla R(r) = \int_{0}^r P(r^*) \text{d}r^* - (1 - \alpha) + \sigma r \\
		& \mathop = \limits^{(a)} \left\{
		\begin{array}{ll}
			\mathrm{CDF}_P(r) - (1 - \alpha) + \sigma r , & 0 \le r \le B , \\
			- (1 - \alpha) + \sigma r, & r < 0 , \\
			\alpha + \sigma r, & r > B .
		\end{array}
		\right.
	\end{aligned}
\end{equation}
where $\mathrm{CDF}_P(r) = \int_0^r P(r^*) \text{d}r^*$ denotes the cumulative distribution function (CDF) of the prior distribution $P(r^*)$, and
(a) follows from the prior distribution $P(r^*)$ being defined over $[0, B]$, while $P(r^*) = 0$ for $r^* \notin [0 , B]$, $\mathrm{CDF}_P(r^*) =1$ for $r > B$, and $\mathrm{CDF}_P(r^*) = 0$ for $r < 0$.

While, the second derivative of $R(r)$ is given by
\begin{equation}
	\begin{aligned}
		\nabla^2 R(r) = \left\{
		\begin{array}{ll}
			P(r) + \sigma , & 0 \le r \le B , \\
			\sigma , & \text{otherwise}.
		\end{array}
		\right.
	\end{aligned}
\end{equation}

Since $\sigma > 0$, we have that $\nabla^2  R(r) = P(r) + \sigma > \sigma$ for all $r\in[0,B]$. 
It follows that $R(r)$ is $\mu$-strongly convex with $\mu \ge  \sigma$.
Similarly, $\nabla^2 R(r) = P(r) + \sigma \le \max_r \{P(r)\} + \sigma$, which implies that $R(r)$ is  $L$-smoothness with $L \leq \max_r \{P(r)\} + \sigma$.

\newpage 
\footnotesize
\bibliographystyle{IEEEtran}
\bibliography{IEEEabrv,./ref.bib}

\newpage

\appendices
\normalsize
\section*{Supplementary Material (Appendix)}

\vspace{-0.25em}
%\subsection{Proof of Lemma \ref{pro:bound}}
\subsection{Proof of Lemma \textcolor{blue}{1}}
\setcounter{equation}{0}
\renewcommand*{\theequation}{A-\arabic{equation}}

Recalling the definition of the mirror map $\mathcal{M}(\cdot)=\nabla R(\cdot)$, the update in the dual space of IM-OCP corresponds to
\begin{align}
	\nabla R(r_t)- \nabla R(r_{t-1})  =- \eta_{t-1} (\alpha - E_{t-1}) \frac{\text{obs}_{t-1}}{p_{t-1}}.
\end{align}
Having assumed the $R(r)$ to be $\mu$-strongly convex, it holds
\begin{equation}
	\begin{aligned}
		\label{eq:first_ineq_sc}
		&  \left\langle \nabla R(r_t) - \nabla R(r_{t-1}) , r_t - r_{t-1} \right\rangle  
		\ge \mu \left\| r_t - r_{t-1} \right\|^2 ,
	\end{aligned}
\end{equation}
\begin{equation}
	\begin{aligned}
		\label{eq:sec_ineq_sc}
		&\left\langle \nabla R(r_t) - \nabla R(r_{t-1}) , r_t - r_{t-1} \right\rangle \leq  \frac{\left\| \nabla R(r_t) - \nabla R(r_{t-1}) \right\|^2}{\mu} 
	\end{aligned}
\end{equation}
Leveraging these two inequalities, we now show by contradiction that the value of the threshold $r_t$ is upper bounded by $B+(1-\alpha)\varpi_{t} \ {\mu}$ for all $t$. Recall that $\varpi_{t} =\max_{i=1,\dots,t-1}\eta_i/p_i$ and hence the sequence $\{\varpi_{t}\}_{t\geq 1}$ is non-decreasing. Assume, without loss of generality, that $t$ is the first instant in which the iterate $r_t>B+(1-\alpha)\varpi_{t} \ {\mu} $. Given that $\varpi_{t} $ is non-decreasing in $t$, it follows that $r_{t}>r_{t-1}$. From inequality \eqref{eq:first_ineq_sc} it follows
\begin{align}
	r_t  \leq r_{t-1}   -\frac{\eta_{t-1} }{\mu}(\alpha - E_{t-1}) \frac{\text{obs}_{t-1}}{p_{t-1}},
\end{align}
from which we conclude that $E_{t-1}=1$ and $\text{obs}_{t-1}=1$ in order for the inequality $r_{t}>r_{t-1}$ be true. At the same time, given that $E_{t-1}=1$ and $\text{obs}_{t-1}=1$, inequality \eqref{eq:sec_ineq_sc} implies
\begin{align}
	r_{t-1} \geq  r_{t} +\frac{\eta_{t-1}(\alpha-1)}{\mu p_{t-1}}>B, 
\end{align}
where the last inequality follows from having initially assumed that $r_t>B+(1-\alpha)\varpi_{t} \ {\mu} $. However, this last inequality leads to a contradiction. In fact, since for $r_{t-1}>B$, we have must have $E_{t-1}=0$ and $\text{obs}_{t-1}=1$.

The lower bound follows similarly.

\vspace{-1em}
%\subsection{Proof of Theorem \ref{pro:mis_IMOCP}}
\subsection{Proof of Theorem \textcolor{blue}{1}}
\setcounter{equation}{0}
\renewcommand*{\theequation}{B-\arabic{equation}}
Note that for IM-OCP it holds that
\begin{align}
	\mathbb{E}[\nabla R(r_{t+1})] &=\mathbb{E}[\nabla R( r_{\tau} )] - \mathbb{E}\left[\sum^T_{t=\tau}\eta_t( E_t - \alpha)\frac{\text{obs}_t}{p_t}\right] \nonumber \\
	&= \mathbb{E}[\nabla R( r_{\tau} )] - \sum^T_{t=\tau}\eta_t( \mathbb{E}[E_t] - \alpha).
	\label{eq:intermediate}
\end{align}

Define $\Delta_1 = 1 / \eta_1$ and $\Delta_i = 1 / \eta_i - 1 / \eta_{i-1}$, and note that $1 / \eta_t = \sum_{i=1}^t \Delta_i$. The expected miscoverage rate at time $T$ is given by
\begin{equation}\label{eq:app5_1}
	\begin{aligned}
		\overline{\mathrm{MisCov}} (T) & = \left| \mathbb{E} \left[  \frac{1}{T} \sum_{i=1}^{T}  \left( \sum_{\tau=1}^{i} \Delta_\tau \right)\eta_{i}(   E_i  -  \alpha)  \right]  \right|  \\
		& =\left| \mathbb{E} \left[ \frac{1}{T}  \sum_{\tau=1}^{T} \Delta_\tau \left(  \sum_{i={\tau}}^{T}  \eta_{i}(  E_i   -  \alpha) \right) \right] \right| \\
		& \le  \frac{\sum_{\tau=1}^T | \Delta_\tau |}{T}  \cdot \left|  \mathbb{E} \left[ \sum_{i=\tau}^{T}  \eta_{i}(  E_i   -  \alpha)   \right] \right|\\
		&\le  \frac{\left\| \Delta_{1:T} \right\|_1}{T}  \cdot \hspace{-1em} \max_{\tau\in \{1, \dots, T\}} \left| \sum_{i=\tau}^{T}  \eta_{i}(  \mathbb{E} \left[  E_i  \right] -  \alpha) \right|
	\end{aligned}
\end{equation}

From \eqref{eq:intermediate}, it holds that
%$\sum_{i=\tau}^{T}  \eta_{i}( \mathbb{E} \left[  E_i \right] -  \alpha)   \hspace{-0.25em} = \mathbb{E} \left[  \nabla R( r_{T+1} ) - \nabla R( r_{\tau} )  \right]$,
\begin{equation}\label{eq:app5_3}
	\begin{aligned}
		\sum_{i=\tau}^{T}  \eta_{i}( \mathbb{E} \left[  E_i \right] -  \alpha)   \hspace{-0.25em} = \mathbb{E} \left[  \nabla R( r_{T+1} ) - \nabla R( r_{\tau} )  \right],
	\end{aligned}
\end{equation}
which allows us to rewrite \eqref{eq:app5_1} as
\begin{equation}
	\label{eq:almost_final}
	\begin{aligned}
		& \overline{\mathrm{MisCov}}(T) \le \frac{\left\| \Delta_{1:T} \right\|_1}{T}  \max_{\tau\in \{1, \dots, T\}}\left|\mathbb{E}\left[  \nabla R( r_{T+1} ) - \nabla R( r_{\tau} )  \right]\right|.
	\end{aligned}
\end{equation}

Since $R(r)$ is an $L$-smooth function, we obtain
\begin{equation}\label{eq:app5_6}
	\begin{aligned}
		& \max_{\tau \in \{1, \dots, T\}}\left| \mathbb{E} \left[\nabla R(r_{T+1}) - \nabla R(r_{\tau}) \right]\right| \\
		&\quad  \le \max_{\tau \in \{1, 2, \dots, T\}} L \left|  \mathbb{E}\left[ r_{T+1} - r_{\tau}\right] \right| L B + \frac{ L \eta_1 }{\mu p_{\min}},
	\end{aligned}
\end{equation}
where the last inequality follows from Lemma 1 and from the assumption of a non-increasing step size sequence $\{\eta_t\}_{t\geq 1}$.

By plugging \eqref{eq:app5_6} into \eqref{eq:almost_final} and noting that $\left\| \Delta_{1:T} \right\|_1 = 1 / \eta_T$, we achieve the final result
\begin{equation}
	\begin{aligned}
		\overline{\mathrm{MisCov}}(T) \le  \frac{1}{T\eta_T} \left( L B + \frac{ L \eta_1 }{\mu p_{\min}}\right) .
	\end{aligned}
\end{equation}

\vspace{-1em}
\subsection{Proof of Theorem \textcolor{blue}{2}}
\setcounter{equation}{0}
\renewcommand*{\theequation}{C-\arabic{equation}}

For any sequence $\{\text{obs}_t\}_{t\geq 1}$ and $u\in \mathbb{R}$, online mirror descent satisfies [\textcolor{blue}{21}, Theorem 6.10] 
%\cite[Theorem 6.10]{orabona2019modern}
%For any sequence $\{\text{obs}_t\}_{t\geq 1}$ and $u\in \mathbb{R}$, online mirror descent satisfies [18, Theorem 6.10]
\begin{equation}
	\begin{aligned}
		& \sum_{t=1}^{T}{ \ell_{1-\alpha} ( r_t , r_t^* ) \frac{\text{obs}_t}{p_t} }  - \sum_{t=1}^{T}{ \ell_{1-\alpha} ( u , r_t^* )   \frac{\text{obs}_t}{p_t} } \\
		& \hspace{1em }\le \frac{D_T}{\eta_T} + \frac{1}{2 \mu} \sum_{t=1}^T \eta_t \left| \nabla \ell_{1 - \alpha} ( r_t , r_t^* ) \frac{\text{obs}_t}{p_t} \right|^2  ,
	\end{aligned}
\end{equation}
where $D_T = \max_{t \in [1 , T]} \mathcal{B}_R ( u , r_t )$.
Taking expectations over the random variables $\{\text{obs}_t\}_{t\geq 1}$
and noting that
\begin{align}
	\mathbb{E} \left[ \left| \nabla \ell_{1 - \alpha} ( r_i , r_i^* ) \frac{\text{obs}_i}{p_i} \right| \right] \leq \mathbb{E} \left[ \left| \frac{\text{obs}_i}{p_i} \right| \right] = 1,
\end{align}
it follows that
\begin{equation}
	\begin{aligned}
		& \overline{\mathrm{Reg}}(T) = \mathbb{E}\left[ \sum_{i=1}^{T}{ \ell_\alpha ( r_i , r_i^* ) \frac{\text{obs}_i}{p_i} } \right] - \sum_{i=1}^{T}{ \ell_\alpha ( u , r_i^* ) }\\
		& \leq \frac{D_T}{\eta_T} + \frac{1}{2 \mu} \sum_{i=1}^T  \frac{\eta_i}{p_i} \le \frac{D_T}{\eta_T} + \frac{1}{2 \mu p_{\min}} \sum_{i=1}^T  \eta_i,
	\end{aligned}
\end{equation}
where $p_{\min} = \min \{ p_i \}$. Setting $u=q_\alpha(r^*_{1:T})$ we obtain the final result.

\vspace{-1em}
\subsection{Proof of Corollary \textcolor{blue}{1}}
\setcounter{equation}{0}
\renewcommand*{\theequation}{D-\arabic{equation}}

For fixed learning rate, by submitting $\eta = cT^{-\beta}$ into miscoverage error rate and regret bound, we have
\begin{equation}
	\begin{aligned}
		\overline{\mathrm{MisCov}}(T) \le & \frac{LB}{c} T^{\beta - 1} + \frac{L}{c \mu p_{\min}} T^{- 1} \sim \mathcal{O} (A T^{\beta - 1}) ,
	\end{aligned}
\end{equation}
and 
\begin{equation}
	\begin{aligned}
		\overline{\mathrm{Reg}}(T) & \le \frac{D_T}{c} T^\beta + \frac{c}{2 \mu p_{\min}} T^{1 - \beta} \sim  \mathcal{O} (T^{\max \{\beta, \beta - 1\}}) .
	\end{aligned}
\end{equation}

Similarly, for decaying learning rate, by submitting $\eta = c t^{-\beta}$ into miscoverage error rate and regret bound, we have
\begin{equation}
	\overline{\mathrm{MisCov}}(T) \le \frac{L B\mu {p_{\min }} + L c}{\mu p_{\min} c}{T^{\beta  - 1}} \sim \mathcal{O}(A T^{\beta - 1} ) ,
\end{equation}
and
\begin{equation}
	\begin{aligned}
		\overline{\mathrm{Reg}}(T) & = \frac{D_T T^\beta}{c} + \frac{c}{2 \mu p_{\min}} \sum_{t=1}^T t^{-\beta}   \sim \mathcal{O} (T^{\max \{\beta, \beta - 1\}}) .
	\end{aligned}
\end{equation}
By setting $\beta =1/2$, i.e., $\eta_t = c \sqrt{t}$, IM-OCP achieves sublinear regret while ensuring the coverage guarantee.

\end{document}